\documentclass[11pt,a4paper]{article}
\usepackage{authblk}
\usepackage[hyperref]{emnlp2020}
\usepackage{times}
\usepackage{latexsym}

\usepackage{subfigure}
\usepackage{microtype}
\usepackage{microtype}
\usepackage{multirow}
\usepackage{makecell}

\usepackage{times}
\usepackage{latexsym}
\usepackage{multirow}
\usepackage{longtable}
\usepackage{supertabular}

\usepackage{graphicx}
\usepackage{verbatim}
\usepackage{caption}
\aclfinalcopy 
\def\aclpaperid{44} 

\title{An Empirical Survey of Unsupervised Text Representation Methods on Twitter Data}
\author[1]{Lili Wang}
\author[2]{Chongyang Gao}
\author[3]{Jason Wei}
\author[4]{Weicheng Ma}
\author[5]{Ruibo Liu}
\author[6]{Soroush Vosoughi}
\affil[1,2,4,5,6] {Department of Computer Science, Dartmouth College}
\affil[3]{ProtagoLabs}
\affil[1,2,4,5]{\texttt{\{first.last.gr\}@dartmouth.edu}}
\affil[3]{\texttt{jason@protagolabs.com}}
\affil[6]{\texttt{soroush.vosoughi@dartmouth.edu}}

\date{}

\begin{document}
\maketitle

\begin{abstract}
The field of NLP has seen unprecedented achievements in recent years. Most notably, with the advent of large-scale pre-trained Transformer-based language models, such as BERT, there has been a noticeable improvement in text representation. It is, however, unclear whether these improvements translate to noisy user-generated text, such as tweets. In this paper, we present an experimental survey of a wide range of well-known text representation techniques for the task of text clustering on noisy Twitter data. Our results indicate that the more advanced models do not necessarily work best on tweets and that more exploration in this area is needed.

\end{abstract}


\section{Introduction}
Recent years have witnessed an exponential increase in the usage of social media platforms. These platforms have become an important part of politics, business, entertainment, and general social life. Correspondingly, the amount of data generated by users on these platforms has also grown exponentially. Though data on social media includes various modalities, such as images, videos, and graphs, text is by far the largest type of data generated by users. Thus, in order to extract knowledge and insight from social media, sophisticated text processing models are needed. Luckily, in parallel to the growth of social media, there has been a rapid rise in the development of sophisticated text representation techniques, the most recent being large-scale pre-trained language models that use Transformer-based architecture \cite{vaswani2017attention}(such as BERT \cite{bert}, and XLNet\cite{xlnet}). These methods can generate general-purpose vector representations of documents that can be used for any downstream task (e.g., sentiment classification). 

However, the representation power of these methods for data from social media is not well understood. This is especially true for tweets which are usually short, noisy, and idiosyncratic. This paper is an attempt to evaluate and catalogue the representation power of a wide range of methods for tweets, starting from very simple bag-of-words representations (or embeddings) to representations generated by recent Transformer-based models, such as BERT. Since we are interested in the general representation power of the methods and not their performance on any specific downstream tasks, we do not fine-tune any of the methods using downstream tasks and use unsupervised evaluation (i.e., clustering) for our survey.

\section{Text Representation Methods}
In this section, we briefly introduce the methods used in our survey, sorted from oldest to newest. For word embedding methods like word2vec, GloVe, and fastText, which dot not explicitly support sentence embeddings, we average the word embeddings to get sentence embeddings. For deep models like ELMo, BERT, ALBERT, and XLNet, we take the average of the hidden state of the last layer on the input sequence axis. Note that some other works use the hidden state of the first token ([CLS]), but in our experiments, we use the pre-trained model without fine-tuning, in this case, the hidden state of [CLS] is not a good sentence representation. Note that we use all these deep neural models without fine-tuning. This is because fine-tuning is usually based on specific downstream tasks which bias the information in the hidden states, weakening the general representation. Note that when we refer to n-gram models we mean models that capture all grams up to and including the n-gram (e.g., bigram models will include bigrams and unigrams).

\smallskip
   \noindent \textbf{1. bag-of-words (BoW).} This is a representation of text that describes the occurrence of words within a document. In our experiments, we use a random sample of 5 million tweets collected from the Internet Archive Twitter dataset \footnote{\url{https://archive.org/search.php?query=collection\%3Atwitterstream&sort=-publicdate}} (IAT) to create a vocabulary. We also remove stop words from the tweets. We try unigram, bigram, and trigram models.  
    
    \noindent \textbf{2. TF-IDF.}. Term frequency–inverse document frequency (TF-IDF) reflects how important a word is with respect to documents in a collection or corpus. We use a similar experimental setup as BoW. 
    
    \noindent \textbf{3. LDA} \cite{lda}. Latent Dirichlet allocation (LDA) is a generative statistical model for capturing the topic distribution of documents in a corpus. We train this model on the IAT dataset. We also remove stop-words and train models with 5, 10, 20, and 100 topics.
    
    \noindent \textbf{4. word2vec} \cite{word2vec}. word2vec is a distributed representation of words based on a model trained on predicting the current word from surrounding context words (CBOW). We train unigram, bigram, and trigram word2vec models using the IAT dataset.
   
    \noindent \textbf{5. doc2vec} \cite{doc2vec}. This model extends word2vec by adding another document vector based on ID. Our model is trained on the IAT dataset.
   
    \noindent \textbf{6. GloVe} \cite{glove}. This model combines global matrix factorization and local context window methods for training distributed representations. We use the 200-dimensional version that was pre-trained on 2 billion tweets.

    \noindent \textbf{7. fastText} \cite{fasttext}. fastText is another word embedding method that extends word2vec by representing each word as an n-gram of characters. We use the 300-dimensional off-the-shelf version which was pre-trained on Wikipedia.
   
    \noindent \textbf{8. Tweet2vec} \cite{tweet2vec}. This model finds vector-space representations of whole tweets by learning complex, non-local dependencies in character sequences. In our experiments, we use the pre-trained best model provided by the authors.\footnote{\url{https://github.com/bdhingra/tweet2vec/tree/master/tweet2vec/best_model} There is another tweet2vec model that uses a  character-level cnn-lstm encoder-decoder \cite{vosoughi2016tweet2vec}, but for the sake of brevity we only show the results for one of the tweet2vec models.}
    
    \noindent \textbf{9. Universal Sentence Encoder (USE)} \cite{use}. USE encodes sentences into high dimensional vectors. The pre-trained encoder comes in two versions, one trained with deep averaging network (DAN) \cite{dan} and one with Transformer. We use the DAN version of USE.

    \noindent \textbf{10. ELMo} \cite{elmo}. This method provides context-dependent word representations based on bidirectional language models. We use the version pre-trained on the One Billion Word Benchmark.
    
    \noindent \textbf{11. BERT} \cite{bert}. BERT is a large-scale Transformer-based language representation model \cite{vaswani2017attention}. We use two off-the-shelf pre-trained versions BERT-base and BERT-large, which are pre-trained on the BooksCorpus and English Wikipedia respectively.
   
    \noindent \textbf{12. ALBERT} \cite{ALBERT}. This is a lite version of BERT, with far fewer parameters. We use two off-the-shelf versions, ALBERT-base and ALBERT-large, which are pre-trained on the BooksCorpus and English Wikipedia respectively.
    
    \noindent \textbf{13. XLNet} \cite{xlnet}. This is an autoregressive Transformer-based language model. Like BERT, XLNet is a large-scale language model with millions of parameters. We use the off-the-shelf versions pre-trained on the BooksCorpus and English Wikipedia.

    \noindent \textbf{14. Sentence-BERT} \cite{sbert}. Sentence-BERT modifies BERT by using siamese and triplet network structures to derive semantically meaningful sentence embeddings. We use five off-the-shelf versions provided by the authors, Sentence-BERT-base, Sentence-BERT-large, Sentence-Distilbert, Sentence-RoBERTa-base, and Sentence-RoBERTa-large, all pre-trained on NLI data.

\section{Experiments}

Since we are interested in measuring the general text representation power of our methods, we use clustering as a way to evaluate the representations generated by each model (instead of any downstream supervised tasks). We use the vector representations of each tweet to run $k$-means clustering for different values of $k$.  We use two tweet datasets for our evaluation. The tweets in these datasets have labels corresponding to their topic which we use as cluster ground-truth for evaluation purposes. 

\smallskip
    \noindent  \textbf{Dataset 1} \cite{zubiaga2014realtime}: This dataset includes 356,782 tweets belonging to 1,036 topics. We use $k\in \{200,400,600,800,1000\}$, for this dataset.  
     
    \noindent  \textbf{Dataset 2} \cite{rosenthal-etal-2017-semeval}: This dataset includes 35,323 tweets belonging to 374 topics. We use $k\in \{100,200,300,400,500\}$, for this dataset.

\subsection{Evaluation Metrics}
 We use a total of six metrics for evaluating the ``goodness" of our clusters, described below. Except for the Silhouette score, all other metrics rely on ground-truth labels.


\smallskip
   \noindent  \textbf{Silhouette score} \cite{rousseeuw1987silhouettes}: A good clustering will produce clusters where the elements inside the same cluster are close to each other and the elements in different clusters are far from each other. The Silhouette score takes both these factors into account. The score goes from -1.0 to 1.0, where higher values mean better clustering.

     \noindent  \textbf{Homogeneity, Completeness, and V-measure}, \cite{rosenberg2007v}: If clusters contain only data points that are members of a single class, in other words, high homogeneity, this usually indicates good clustering. Similarly, if all members of a given class are assigned to the same cluster, in other words, high completeness, this usually indicates good clustering. The Homogeneity and Completeness scores are between 0.0 and 1.0, where higher values correspond to better clustering. The V-measure score is the harmonic mean of Homogeneity and Completeness.

   \noindent  \textbf{Adjusted Rand Index (ARI)} \cite{hubert1985}: The Rand Index can be used to compute the similarity between generated clusters and ground-truth labels. This is done by considering all pairs of samples and seeing whether their label agreement (i.e., belonging to the same ground-truth cluster or not) matches the generated cluster agreement (i.e., belonging to the same generated cluster or not). The raw RI score is then “adjusted for chance” into the ARI. score using the following formula: 
    The ARI score can be between -1.0 and 1.0, where random clusterings have an ARI close to 0.0 and 1.0 stands for perfect clustering.

\begin{figure}[h]
    \centering 
    \includegraphics[width=.95\linewidth]{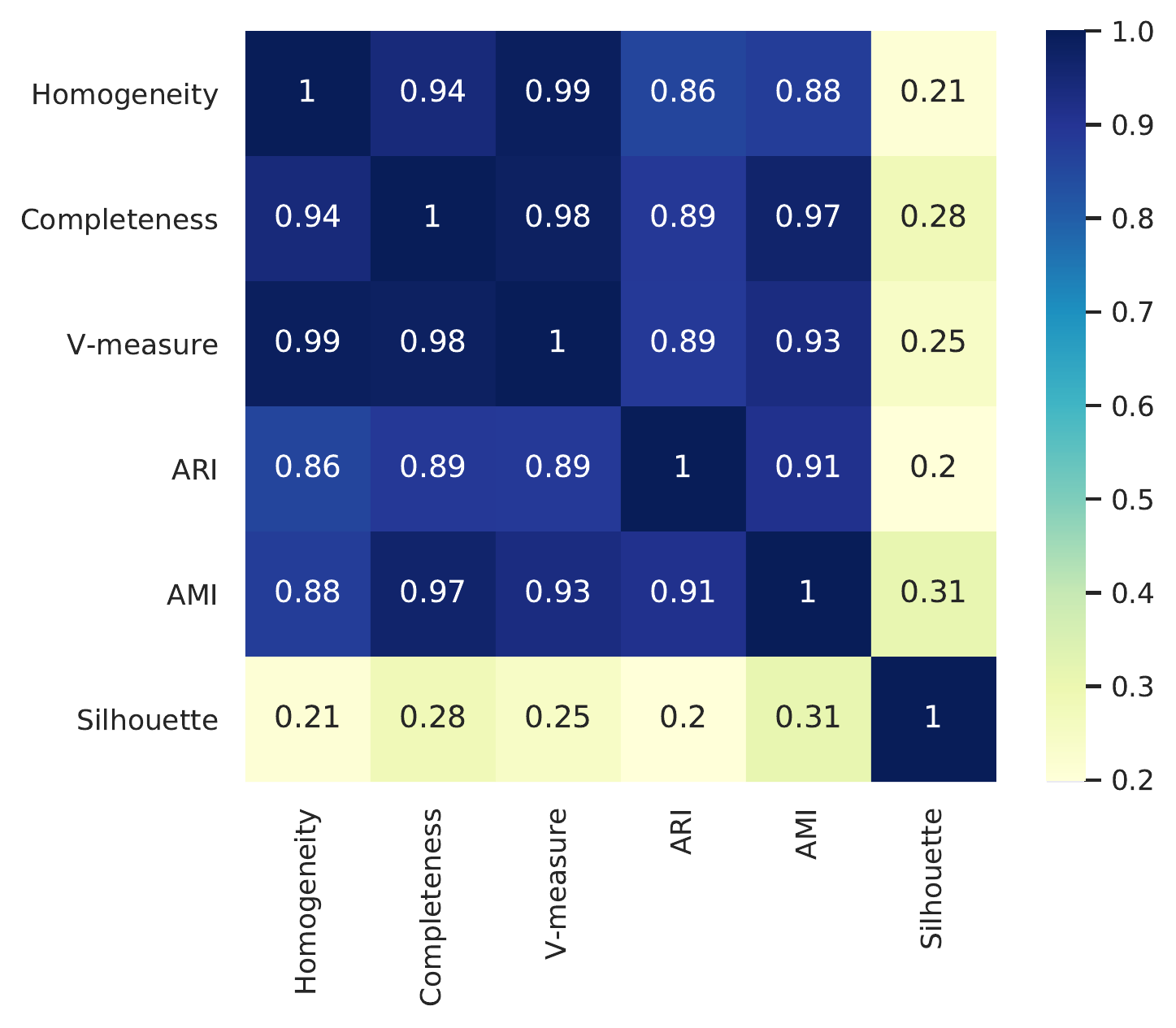}
    \caption{Confusion matrix of the correlation (Pearson's r) between each pair of methods.}
    \label{fig:pearson}
\end{figure}

 \noindent    \textbf{Adjusted Mutual Information (AMI)} \cite{vinh2010information}: The Mutual Information (MI) score is an information-theoretic metric that measures the amount of "shared information" between two clusterings. The Adjusted Mutual Information (AMI) is an adjustment of the Mutual Information (MI) score to account for chance.  It accounts for the fact that the MI is generally higher for two clusterings with a larger number of clusters, regardless of whether there is actually more information shared. 

    The AMI score can be between 0.0 and 1.0, where random clusterings have an AMI close to 0.0 and 1.0 stands for perfect clustering.

\section{Results \& Discussion}
\begin{figure*}[ht]
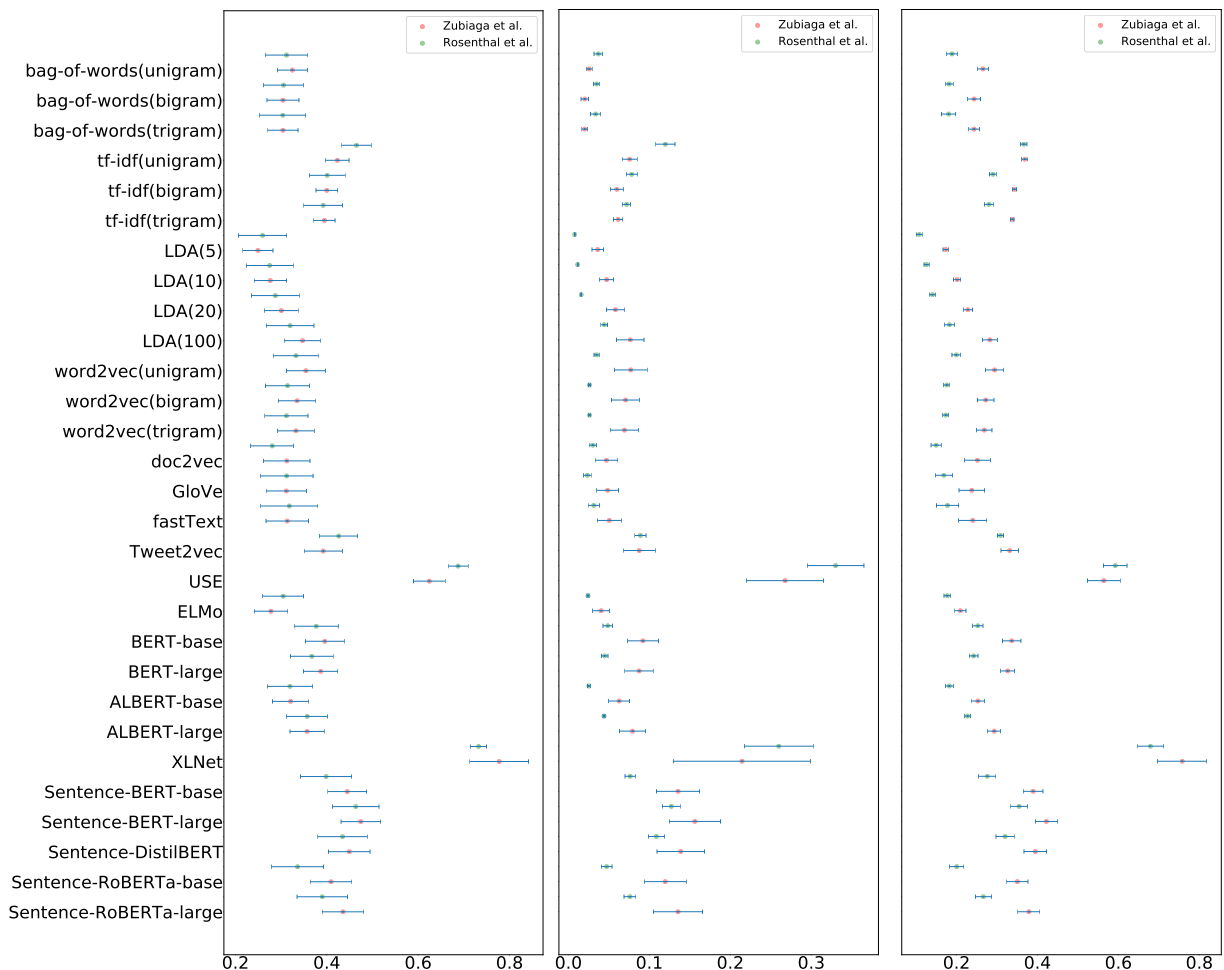
 
\subfigure{
\begin{minipage}[t]{0.4414\linewidth} 
\centering
\includegraphics[width=\linewidth]{V_measure.pdf}
\end{minipage}%
}%
\subfigure{
\begin{minipage}[t]{0.2699\linewidth} 
\centering
\includegraphics[width=\linewidth]{ARI.pdf}
\end{minipage}
}%
\subfigure{
\begin{minipage}[t]{0.2699\linewidth} 
\centering
\includegraphics[width=\linewidth]{AMI.pdf}
\end{minipage}
}%
\centering
\caption{The V-measure (left), ARI  (middle), and AMI (right) of all the methods on the two datasets. The points in the figure denote the average value across different $k$ values and the blue lines denote the standard deviations. The methods are sorted from the oldest to the newest. }
\label{fig:cluster}
\end{figure*}

\begin{figure*}[ht]
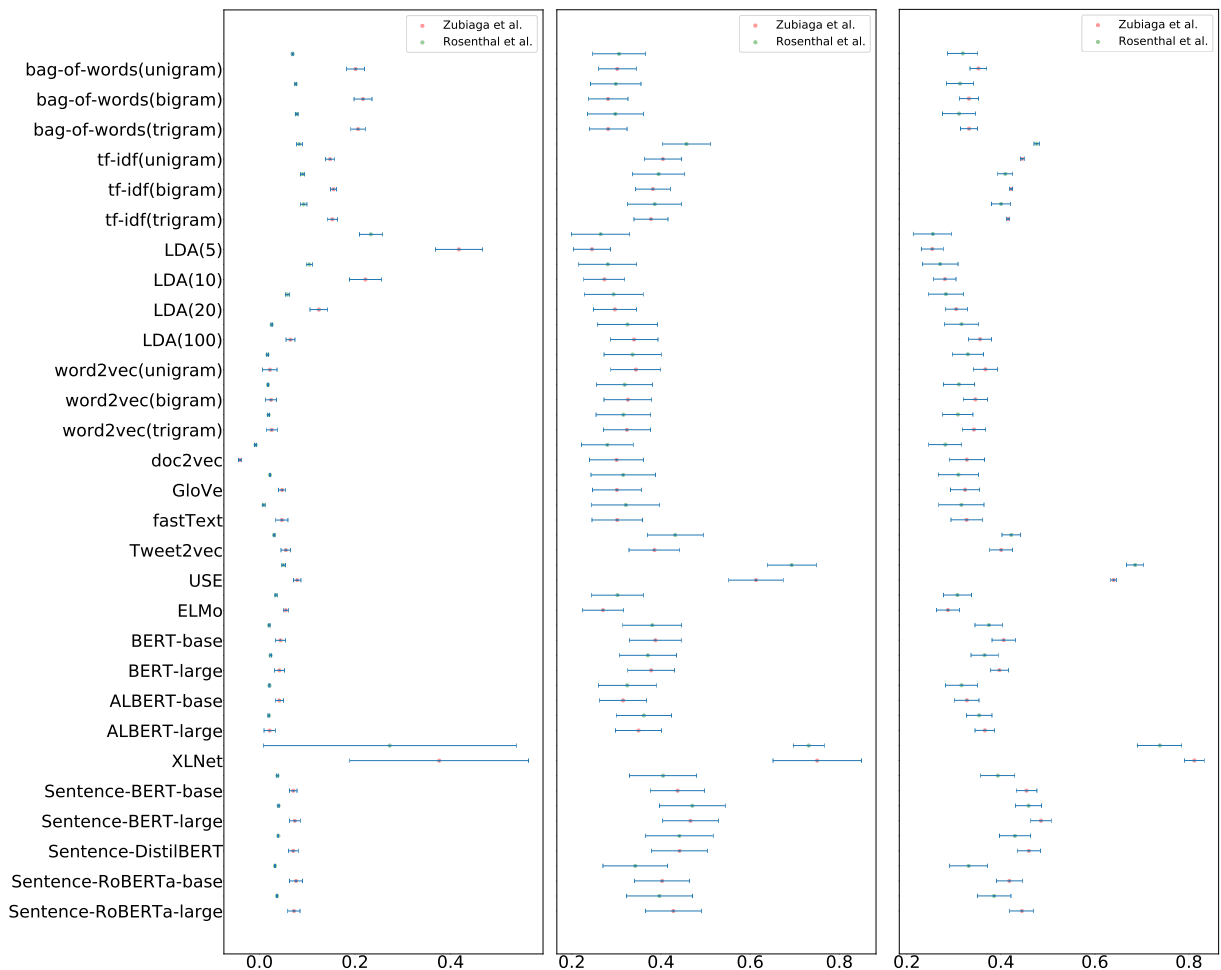
 
\subfigure{
\begin{minipage}[t]{0.4414\linewidth} 
\centering
\includegraphics[width=\linewidth]{Silhouette.pdf}
\end{minipage}%
}%
\subfigure{
\begin{minipage}[t]{0.2679\linewidth} 
\centering
\includegraphics[width=\linewidth]{Homogeneity.pdf}
\end{minipage}
}%
\subfigure{
\begin{minipage}[t]{0.2696\linewidth} 
\centering
\includegraphics[width=\linewidth]{Completeness.pdf}
\end{minipage}
}%
\centering
\caption{The Silhouette (left), Homogeneity (middle), and Completeness (right) of all the methods on the two datasets. The points in the figure denote the average value across different $k$ values and the blue lines denote the standard deviations. The methods are sorted from the oldest to the newest. }
\label{fig:cluster2}
\end{figure*}

For each dataset, we average the scores from $k$-means clustering with different values of $k$. Though we use several metrics in our evaluations for the sake of being thorough, most of the metrics are in fact highly correlated. Fig. \ref{fig:pearson} shows the correlation between each pair of metrics (calculated based on the clustering results of our methods). We can see that all the \emph{external} evaluation metrics (Homogeneity, Completeness, V-measure, AMI, and ARI, which need external ground-truth labels) highly agree with each other while the \emph{internal} evaluation metric (Silhouette score, which does not need external ground-truth labels) does not.

The clustering results are shown in Fig. \ref{fig:cluster} and Fig. \ref{fig:cluster2}, the methods in both figures are sorted based on the date of their release to capture the advancements in NLP. Unlike conventional tasks and datasets (such as the GLUE benchmark \cite{wang2018glue}), there does not seem to be a very clear trend of improvement for capturing tweet representations. The more advanced models are not necessarily the best. Notably, the BERT family of large-scale pre-trained language models (ALBERT, Sentence-BERT, etc) do not vastly or consistently outperform much simpler methods such as bag-of-words and tf-idf. XLNet, on the other hand, seems to be the best performing method for capturing tweet representations, followed closely by USE. Interestingly, XLNet is also the most volatile with respect to the choice of $k$ in our clustering. We think XLNet outperforms other comparable (in terms of complexity) models such as BERT since it uses permutation language modeling, allowing for prediction of tokens in random order. This might make it more robust to the noisy user-generated text, such as tweets. We think that our results are unexpected and inconclusive, demonstrating that much is still unknown about the performance of the most recent models on noisy and idiosyncratic user-generated text.

Very recently, a large-scale pre-trained BERT model for English Tweets was trained and released \cite{nguyen2020bertweet}. This model was released just days before the publication of this paper and thus we did not have time to thoroughly compare its performance against the other models. However, we believe this model is a step in the right direction as we have shown in this paper that models trained on standard English corpora do not perform well on Tweets.

\section{Conclusion}
In this paper, we presented an experimental survey of 14 methods for representing noisy user-generated text prevalent in tweets. These methods ranged from very simple bag-of-words representations to complex pre-trained language models with millions of parameters. Through clustering experiments, we showed that the advances in NLP do not necessarily translate to better representation of tweet data. 

We believe more work is needed to better understand and potentially improve the performance of the more recent methods, such as BERT, on noisy, user-generated data.

\bibliography{emnlp2020}
\bibliographystyle{acl_natbib}

\end{document}